\documentclass[runningheads]{llncs}
\usepackage{graphicx}

\usepackage{tikz}
\usepackage{comment}
\usepackage{amsmath,amssymb} 
\usepackage{color}
\usepackage{soul}
\usepackage{url}
\usepackage[hidelinks]{hyperref}
\usepackage[utf8]{inputenc}
\usepackage{amsmath}
\usepackage{booktabs}
\usepackage{algorithm}
\usepackage{amssymb}
\usepackage{enumitem}
\usepackage{algpseudocode}
\usepackage{multirow}
\usepackage{makecell}
\usepackage{subfigure}
\usepackage{color,wrapfig}
\usepackage{orcidlink}
\usepackage[accsupp]{axessibility}  

\begin{document}
\pagestyle{headings}
\mainmatter
\def\ECCVSubNumber{1029}  

\title{Few-Shot Class-Incremental Learning via Entropy-Regularized Data-Free Replay} 

\titlerunning{FSCIL via Entropy-Regularized Data-Free Replay}
%
\author{Huan Liu\inst{1,2}\orcidlink{0000-0001-7155-3663} \and
Li Gu\inst{1}\orcidlink{0000-0002-4447-4967} \and
Zhixiang Chi\inst{1}\orcidlink{0000-0003-4560-4986}\and
Yang Wang\inst{1,3}\orcidlink{0000-0001-9447-1791}
\and
Yuanhao Yu\inst{1}\orcidlink{0000-0001-8176-9716}
\and
Jun Chen\inst{2}\orcidlink{0000-0002-8084-9332}
\and
Jin Tang\inst{1}
}
\authorrunning{H. Liu et al.}
%
\institute{Noah’s Ark Lab, Huawei Technologies \and McMaster University, Canada
\and University of Manitoba, Canada
\\
\email{\{liuh127, chenjun\}@mcmaster.ca},
\email{\{li.gu, zhixiang.chi, yang.wang3, yuanhao.yu, tangjin\}@huawei.com}}
\maketitle

\begin{abstract}

Few-shot class-incremental learning (FSCIL) has been proposed aiming to enable a deep learning system to incrementally learn new classes with limited data. Recently, a pioneer claims that the commonly used replay-based method in class-incremental learning (CIL) is ineffective and thus not preferred for FSCIL. 
This has, if truth, a significant influence on the fields of FSCIL. 
In this paper, we show through empirical results that adopting the data replay is surprisingly favorable. However, storing and replaying old data can lead to a privacy concern. To address this issue, we alternatively propose using data-free replay that can synthesize data by a generator without accessing real data. In observing the the effectiveness of uncertain data for knowledge distillation, we impose entropy regularization in the generator training to encourage more uncertain examples. Moreover, we propose to relabel the generated data with one-hot-like labels. This modification allows the network to learn by solely minimizing the cross-entropy loss, which mitigates the problem of balancing different objectives in the conventional knowledge distillation approach. Finally,  we show extensive experimental results and analysis on CIFAR-100, miniImageNet and CUB-200 to demonstrate the effectiveness of our proposed one.
  

\end{abstract}

\section{Introduction}

Recently, there has been a tremendous success in using deep learning technologies
\cite{lecun2015deep} 
in large-scale image recognition tasks. Despite the remarkable success, they usually train a neural network to learn a mapping on a large amount of data. The model is then fixed and cannot be changed according to the users' needs. In contrast, humans can continually learn new knowledge throughout their lifetime. Inspired by this human capability, class-incremental learning (CIL) has been introduced to allow the neural network to continually update after new classes or environments are encountered. Despite the practical value of CIL, it usually suffers severely from the well-known catastrophic forgetting issue \cite{french1999catastrophic}, 
especially when the old model is fine-tuned only with a large amount of new data. CIL assumes that we have enough training data for each new class. In real-world applications, this assumption is not practical since it is expensive to collect a large number of examples for each new class. In this paper, we consider the more realistic setting, i.e., few-shot class incremental learning (FSCIL). FSCIL aims to design learning systems that can incrementally learn new classes with limited data. This problem is more challenging than CIL since a system can easily overfit very few new examples and severely forget the old knowledge.

\begin{wrapfigure}{l}{7cm}
    \centering
    \includegraphics[width=1.0\linewidth]{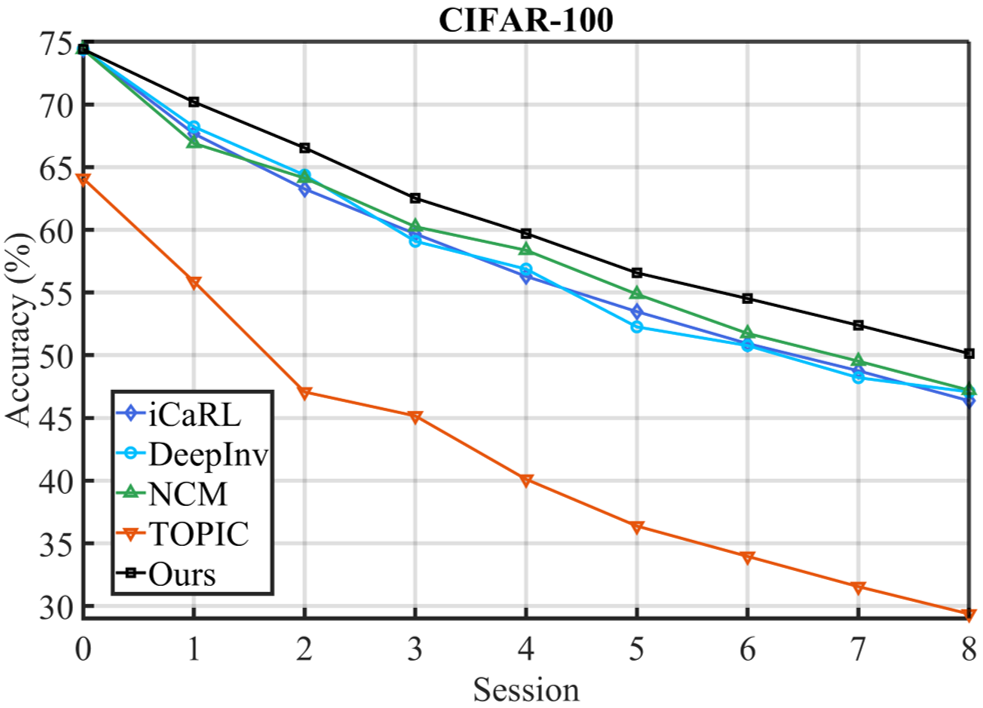}
    \caption{ The comparison between TOPIC \cite{topic} and three replay-based methods, i.e. iCaRL \cite{icarl}, NCM\cite{ncm} and DeepInversion \cite{yin2020dreaming}. Our approach is also included for reference.}
    \label{fig:illustration}
\end{wrapfigure}
A naive way to address the FSCIL problem is to directly apply the commonly used approaches in the CIL, such as data replay \cite{icarl}. Current literature demonstrates several ways for addressing the FSCIL problem, but none of them attempts to apply the replay-based method. One possible reason is that the pioneer \cite{topic} emphasizes the defectiveness of adopting data replay in addressing FSCIL. The authors denote that vanilla data replay can cause a significant problem of imbalance \cite{ncm} and thus is not preferred in FSCIL as new classes are learned with very limited data. Extensive experiments also shows that several replay-based CIL approaches performs extremely bad in FSCIL.
The strong conclusion might prevent the researchers from exploring the line of approaches further. Intuitively, data replay is conceptually suitable for addressing the kind of incremental learning problem. To confirm our intuition, we re-implement several replay-based CIL methods under the setting of FSCIL and carefully tune their performance to the best. Somewhat surprisingly, we find through experimental results that the affirmation of the TOPIC \cite{topic} is \textbf{\textit{not true.}} 
Figure \ref{fig:illustration} presents the comparison between TOPIC and replay-based approaches, from which we can observe that all the replay-based approaches outperform TOPIC \cite{topic} by a significant margin. This evidence attracts our interest in exploring a specific design for FSCIL using data replay.
However, another concern might arise as the replay of previous real data is not permitted in many computer vision applications since it can violate legality concerns.

Recently, DeepInversion \cite{yin2020dreaming} has been proposed in addressing CIL without violating the legality issues. It learns to optimize random noises to photo-realistic images by inverting a reference network without accessing the real data. 
However, this approach leverages a constraint that generated images should strictly belong to a particular class with high confidence. 
In FSCIL, since the incremental classes only provide limited data examples, the network's predictions of these few-shot learned classes are usually uncertain.
Therefore, it can be expected that only a small collection of images that represent part of the class identities can be generated for replay using \cite{yin2020dreaming}. As a result, DeepInversion is not suitable for FSCIL.

In this paper, we propose a replay-based method in FSCIL under the data-free setting. In observing the fact that distilling knowledge using uncertain data is more effective since they are usually close to the model's decision boundaries \cite{belief}, we introduce an entropy-regularized method to explicitly encourage the replayed data to be close to decision boundaries given by the reference model. This is achieved by maximizing the information entropy during the training of the generator for data replay. This allows us to produce more uncertain data for effective knowledge distillation to mitigate forgetting. 
However, we find in our experiments that using vanilla knowledge distillation in FSCIL is highly non-trivial, which is consistent with the analysis in TOPIC \cite{topic}. Even though we can carefully tune the current replay-based method to achieve satisfying performance (as is shown in Figure \ref{fig:illustration}), tuning the hyper-parameters is a cumbersome task in general. For example, the hyper-parameter to weight for cross-entropy loss and knowledge distillation loss should be precisely selected. Because there is a dilemma to balance between the contribution of cross-entropy loss and KL divergence loss when using replayed data to distill knowledge from an old model to a new one.
Specifically, the learning rate required to minimize cross-entropy loss is usually large, while a large learning rate can potentially cause instability when minimizing KL divergence.
To address this issue, we propose re-labelling the generated data samples by one-hot-like labels using the old model and adding the generated data pairs together with novel real data to the current dataset. Then, we can solely adopt the cross-entropy loss to alleviate forgetting and simultaneously learn new classes.

In summary, we have the following contributions:
\vspace{-0.6em}
\begin{itemize}
\item Through experiments, we point out a misleading conclusion in the current literature that data replay is not preferred for FSCIL. Our re-implementation of several replay-based approaches demonstrates that data replay is actually effective. 
\item We propose to introduce data-free replay in handling FSCIL problem. Thanks to the nature of data-free replay, we can replay observed data without violating legality concerns.

\item We improve the current data-free replay method by introducing an entropy term to penalize the generator's training and introduce re-label generated data to avoid the problem of balancing between different loss functions in the typical knowledge distillation. Our method demonstrates that it is possible and even preferred to adopt a replay-based method in FSCIL.
\item Extensive experiments show that we achieve state-of-the-art performance for the FSCIL setting.
\end{itemize}

\section{Related Works}

\subsection{Class-Incremental Learning}
Recently, extensive attention has been attracted to enabling artificial intelligent systems to learn incrementally. 
In this paper, we focus on methods that address the class-incremental learning problem. iCaRL \cite{icarl} incrementally learn nearest-neighbor classifier to predict novel classes and store real data of old classes against forgetting via knowledge distillation. However, this approach violates data privacy. EEIL \cite{eeil} design an end-to-end learning system for incremental learning, where a cross-entropy loss and knowledge distillation loss are respectively adopted to learn novel classes to preserve old knowledge. NCM \cite{ncm} introduces cosine normalization to balance between the classifier for previous and novel data. 
Besides, inspired by the remarkable progress in self-supervised visual representation learning \cite{rep1,rep2,rep3,rep4,liang2022self}, \cite{fini2022self} propose to enable continual learning without labeled data.
In this paper, we aim to address a more realistic and challenging problem, i.e., few-shot class incremental learning (FSCIL). Unlike the typical class-incremental learning with numerous data for new classes' training, there are limited samples to be provided for training new classes under FSCIL. 

\subsection{Few-shot Class-incremental Learning}

The problem of few-shot class-incremental learning (FSCIL) is firstly introduced in TOPIC \cite{topic}. It proposes a single neural gas (NG) network to stabilize feature typologies for observed classes and implement to grow NG to adapt to new training data. Recently, many works have been proposed aiming to address this problem.  \cite{zhu2021self} propose a self-promoted prototype refinement mechanism to explicitly consider the dependencies among classes, which results in an extensible feature representation. To further leverage the benefits from re-projected features,  a dynamic relation projection module is designed to update prototypes using relational metrics. \cite{cheraghian2021synthesized} introduce a mixture of subspace-based method with synthetic feature generation, where the catastrophic forgetting problem is handled using a mixture of subspace and synthetic feature generation can help alleviate the over-fitting problem of novel classes. 
\cite{cec} introduce a pseudo incremental learning paradigm incorporating meta-learning approach to enable the graph network to update the classifier according to the global context of all classes. \cite{metafscil} proposes a bi-level optimization technique to learn how to incrementally learn in the setting of FSCIL.
Despite the success of these approaches, we notice that the replay-based method is under-studied.

\subsection{Data-free Knowledge Distillation}
Data-free replay has been proven effective in dealing with the incremental learning problem. There are two lines of works that synthesize data of old classes. The first line of research \cite{gan_memory,cl_gr,shankarampeta2021few} requires a generator to be trained on the original image samples. 
However, this setting is not preferred as the generator must be stored and transmitted to the next training session. 
Our method is proposed following the second line of generative replay that only uses a trained network as a reference to synthesize pseudo images. DeepInversion \cite{yin2020dreaming} propose to 'invert' an already-trained network to synthesize images of particular classes starting from random noises. \cite{Data-free-student,belief} alternatively uses a GAN architecture to synthesize images, where they fix a trained network as a discriminator and optimize a generator to derive images that can be adopted to distill knowledge from the fixed network to a new network.
Recently, \cite{smith2021always,yin2020dreaming,xin2021memory} have integrated the idea of generative reply into addressing class-incremental learning problem using Deep-Inversion. However, due to the more challenging setting of FSCIL,  these approaches cannot be simply migrated to handle the few-shot scenario.

\section{Preliminaries}
In this section, we formally describe the problem setting of few-shot class-incremental learning (FSCIL). We also introduce data-free replay, which is the basis of our approach. 

\subsection{Problem Setting}
FSCIL aims to enable a learning system to continually learn novel classes from very few data examples. 
The problem is defined as follows. Let $\{\mathcal{D}_{train}^0, \mathcal{D}_{train}^1, $ $..., \mathcal{D}_{train}^N \}$ and $\{\mathcal{D}_{test}^0, \mathcal{D}_{test}^1, ..., \mathcal{D}_{test}^N \}$ respectively denote the collection of training datasets and testing datasets, where N is the total number of learning sessions. The class labels of each session are disjoint. 
Following the setting in \cite{topic}, $\mathcal{D}_{train}^0$ is the base training dataset. The base training set contains a large number of classes where each class has enough training samples. For each subsequent session $i$ ($i=1,2,...,N$), the corresponding training set $\mathcal{D}_{train}^i$ only contains a small number of classes where each class only has very few training examples. At the $i$th session, only $\mathcal{D}_{train}^i$ can be accessed for training. After the $i$-th session, the model is evaluated on its performance in recognizing all object classes that have appeared so far. In other words, the test dataset $\mathcal{D}_{test}^i$ contains examples of all object classes that have appeared in $\{\mathcal{D}_{train}^0, \mathcal{D}_{train}^1, ..., \mathcal{D}_{train}^i\}$.

\subsection{Data-free Replay}
Here, we briefly show the procedures of training a generator for synthesizing critical samples of observed classes. 

Given a trained model $\mathcal{T}(\cdot;\theta)$, our goal is to train a generator $\mathcal{G}(\cdot;\theta_{G})$ that can synthesize critical samples for replay purpose. We follow \cite{belief} to train the generator by including an auxiliary model $\mathcal{A}(\cdot;\theta_{A})$ as a helper to assist the convergence of the generator. Specifically, there are two phases in training the generator, i.e., the knowledge transferring phase and the generator evolving phase.
In the knowledge transferring phase, the generator takes a noise vector $z \sim \mathcal{N}(0, I)$ and outputs a generated image $x$. Then, each sample $x$ is fed into the original model $\mathcal{T}(\cdot;\theta)$ and the auxiliary model $\mathcal{A}(\cdot;\theta_{A})$, where the input sample is mapped to
logit $o$ and $o_A$ (i.e., inputs of softmax function), respectively. We optimize on the auxiliary model to let its outputs match the original model using the following loss:
\begin{equation}
    \mathcal{L}_A = ||\mathcal{T}(\mathcal{G}(z)) - \mathcal{A}(\mathcal{G}(z))||^2_2
\end{equation}
Here, the primary purpose of this update is to enable the auxiliary model to be close to the original model. Then we conduct the generator evolving phase. In this phase, the goal is to optimize the generator so that it can produce more critical samples for knowledge transfer.
To achieve this, we optimize the generator using the following loss: 
\begin{equation}\label{eq:G}
    \mathcal{L}_G = -||\mathcal{T}(\mathcal{G}(z)) - \mathcal{A}(\mathcal{G}(z))||^2_2
\end{equation}

By maximizing the distance between the outputs of the old model and the auxiliary model, we push the generator to produce samples that are hard to be learned by the auxiliary model. We alternate between the knowledge transferring phase and the generator evolving phase. The generator can finally produce more critical examples for transferring knowledge. Note that we adopt mean square error (MSE) as objective in both phases other than KL divergence adopted in \cite{belief}, because logit matching has better generalization capacity \cite{kl_mse}.

In \cite{belief}, the authors also indicate that the uncertain samples (i.e. those with less confident predictions) are usually close to the boundary decisions of the original model. This property has important implication in FSCIL as models learned with few-shot examples often assign low confidence to an input image. By observing that, we propose using information entropy to quantify the confidence and impose an entropy regularization to explicitly encourage the generator to produce more uncertain data. 

\begin{figure}[!t]
    \centering
    \includegraphics[width=0.9\linewidth]{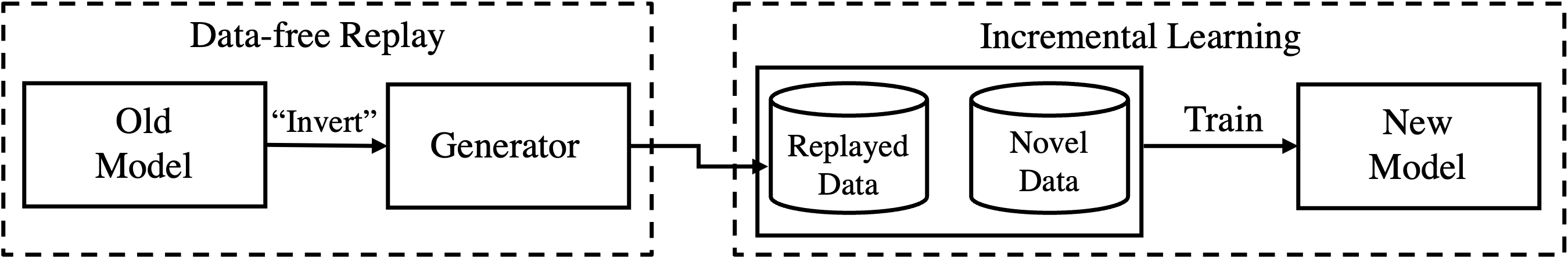}
    \caption{Overview of our system. We first train a generator to synthesize old samples given an old model. Then, the replayed data together with novel data are used to train a new model.}
    \label{fig:workflow}
\end{figure}

\section{Methodology}
In this section, we present our data-free replay approach in addressing the FSCIL problem. Figure~\ref{fig:workflow} shows an overview of our approach at a particular session $i$. Given the old model from the previous session, we first ``invert'' the old model to obtain a generator. The generator is used to synthesize examples of classes that have appeared in the previous sessions $\{0,1,...,i-1\}$. These synthetic examples will be used for replay in order to alleviate the catastrophic forgetting issue. The training at the $i$-th session is performed using both those synthetic examples and the training examples for new classes in this session. However, we have found that a naive application of data-free replay does not work well since the synthesized examples are often far away from any decision boundaries. As a result, they do not have much influence on the learned model. Instead, we introduce entropy-regularized data-free replay to explicitly encourage the generator to synthesize examples that are close to decision boundaries. Then, we show our incremental learning algorithm that uses the replayed data samples.

\begin{figure}[!t]
    \centering
    \includegraphics[width = 1.0\linewidth]{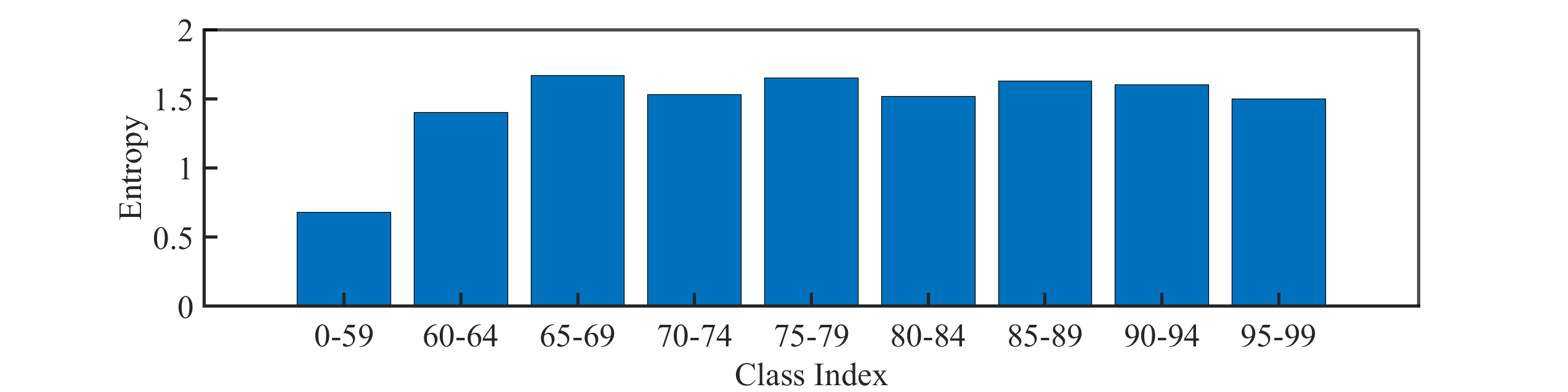}
    \caption{The average entropy of network outputs in different classes. Classes 0-59 are the base classes, where all data samples are available for training. Classes 60-99 are the incrementally learned classes, where each class has 5 training images. The few-shot learned classes (60-99) have higher entropy than the regularly trained classes (0-59). The experiment is conducted on CIFAR-100. }
    \label{fig:entropy}
\end{figure}

\subsection{Entropy-regularized Data-free Replay} \label{sec:entropy}

Information entropy is a well-defined measurement for uncertainty. High entropy denotes low confidence and vise versa. In a few-shot incrementally trained model, the high entropy response of input usually can be identified as the case that the input is on its decision boundary \cite{belief} or is learned in a few-shot incremental session (i.e. not in base classes). To demonstrate this phenomenon, we show the output entropy of a continually learned model \cite{cec} in Figure \ref{fig:entropy}. We can observe that the classes trained using a large amount of data (0-59) maintain a low entropy, while all few-shot learned classes (60-99) have a high entropy on test images. This indicates that the network cannot assign confident predictions to images of few-shot learned classes. Motivated by this observation, we propose an entropy regularization to guide the generator to synthesize uncertain images. 
Specifically, we optimize the generator to maximize the entropy of predictions from the original model.
we measure the Shannon entropy of each prediction using:
\begin{equation}\label{eq:entropy}
    H(\hat{y}) = -\sum_c p(o^c)\log p(o^c)
\end{equation}
\noindent
where $c$ denote the class index and $p(o^c)$ represents the probability. Since our objective is to maximize the entropy of the teacher's prediction to the generated image, during the generator evolving phase, Eq. \ref{eq:G} becomes:
\begin{equation}\label{eq:G_star}
    \mathcal{L}_G^* = -||\mathcal{T}(\mathcal{G}(z)) - \mathcal{A}(\mathcal{G}(z))||_1 - H(\mathcal{T}(\mathcal{G}(z)))
\end{equation}

We use the toy experiment proposed by \cite{belief} to illustrate the effects of our entropy regularization in Figure \ref{fig:toy}. Thanks to the proposed entropy regularization, we can observe that the generated samples (yellow crosses) are more likely to lie on the high-entropy regions (decision boundaries).

\begin{figure}[t]
    \centering
    \subfigure[]{%
        \includegraphics[width=0.3\linewidth]{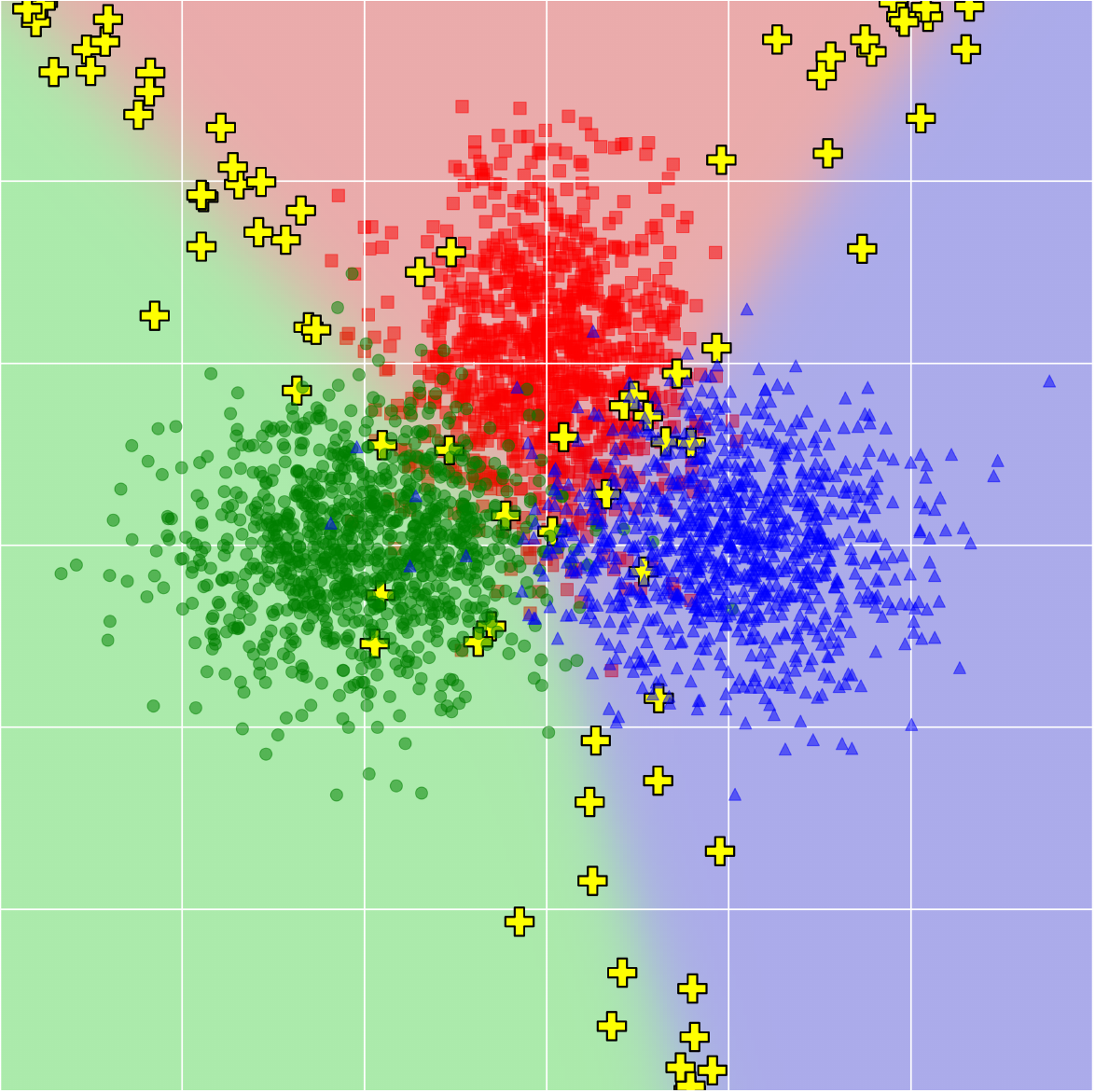}
        \label{fig:toy_wo_entropy}
    } 
    \subfigure[]{%
        \includegraphics[width=0.3\linewidth]{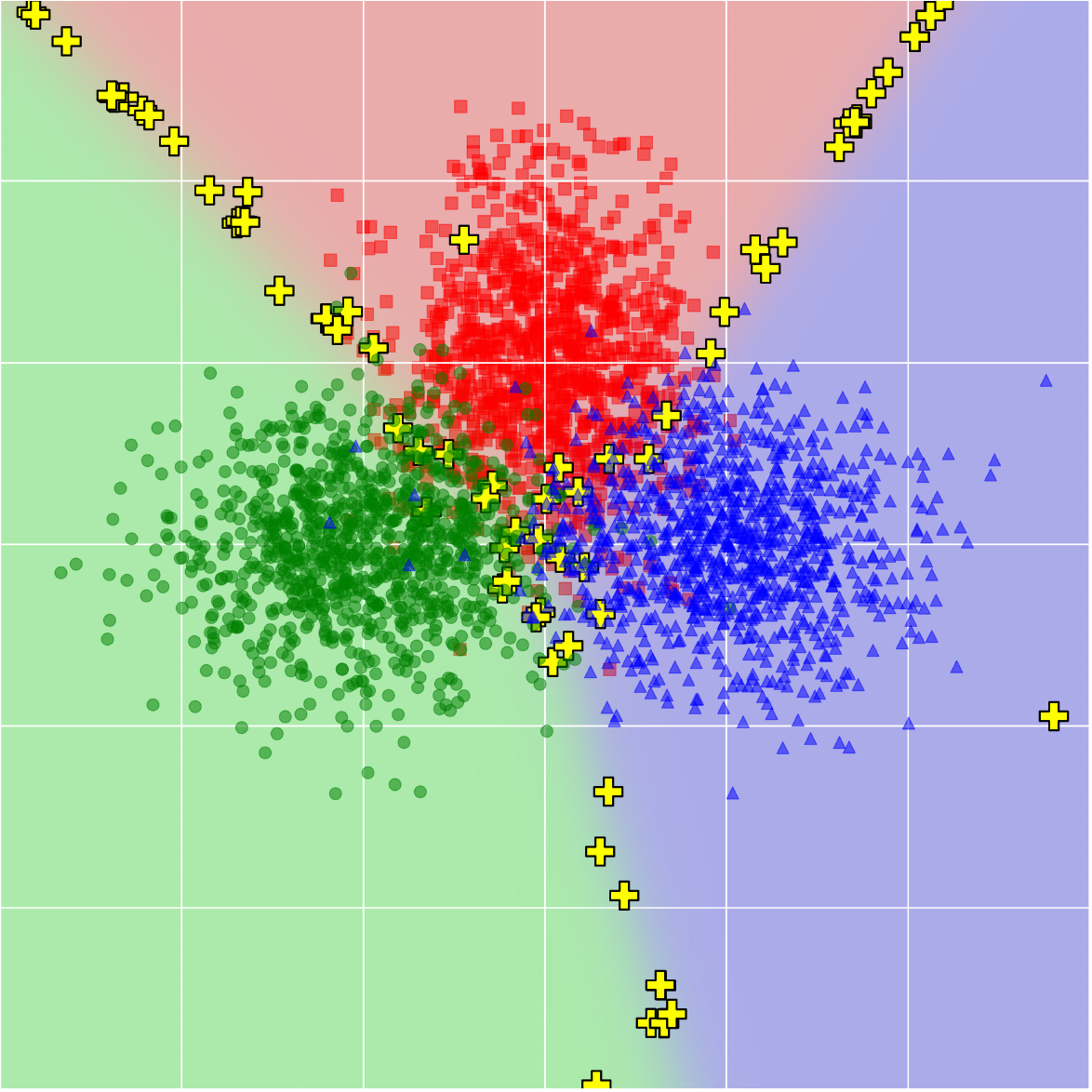}
        \label{fig:toy_w_entropy}
    } 
	\caption{Toy example to illustrate the generated data (yellow crosses). \subref{fig:toy_wo_entropy} shows the data generated using a normal generator. \subref{fig:toy_w_entropy} shows the case when the generator is trained with entropy constraint. Red, green, and blue points are the real data. The background shows the decision boundaries of the model trained on real data. }
	\label{fig:toy}
\end{figure}

\subsection{Learning Incrementally with Uncertain Data} \label{sec:incremental}

Given an old model learned on previous session $i-1$ with parameters $\theta_{i-1}$, our goal is to add linear classifier nodes parameterized by $\theta^{l^*}_i$ for new session and update the evolved model $\mathcal{T}_i(\cdot;\theta_i)$ using both generated data and novel data. Note that a new model on current session $i$ is initialized by the following parameters:
\begin{equation}
    \theta_i = \{ \theta^b_{i-1}, \theta^{l}_{i-1}, \theta^{l^*}_i \}
\end{equation}
where $\theta^b_{i-1}$ and $ \theta^{l}_{i-1}$ are respectively the parameters of old backbone and old linear classifier, i.e., $\theta_{i-1} = \{ \theta^b_{i-1}, \theta^{l}_{i-1} \}$. $\theta^{l^*}_i$ is randomly initialized.

Intuitively, with the old data samples replayed by our generator, we can go against forgetting in incremental learning by distilling the knowledge from the old model to a new one. A simple way to achieve this is that we can let a new model imitate the output of the old one given generated data following the vanilla knowledge distillation pipeline. 
However, it is non-trivial to adopt the knowledge distillation method in addressing the FSCIL problem directly. 
Although we show in Figure \ref{fig:illustration} that using vanilla knowledge distillation with replayed data can achieves good performance, we also find in the process of re-implementation that the performance of current replay-based approaches is very sensitive to the selection of hyper-parameters. A careless tuning of these replay-based methods can easily result in the opposite conclusion (TOPIC \cite{topic} denotes that data replay is not preferred.). For example, 
since the cross-entropy loss and knowledge distillation loss are less balanced and require careful hyper-parameter tuning, it makes the training process less stable, especially under the few-shot scenario.
To alleviate the burden, we introduce to re-label the generated data with one-hot-like labels. Then, we can use the generated data together with the novel data to train an evolved network by minimizing cross-entropy solely. Specifically, unlike the vanilla knowledge distillation, we alternatively assign hard labels to the generated images and eliminate the KL divergence loss. Given a synthetic image $x^*$ and its pseudo label $\mathcal{T}_{i-1}(x^*)$ produced by the old model, we assign a one-hot label $y^* \in \{0, 1, 2, . . . , C-1\}$ to $x^*$ using the following equation, where $C$ is the total number of classes. 
\begin{equation}\label{eq:arg}
    y^* \leftarrow argmax (\mathcal{T}_{i-1}(x^*))
\end{equation}

The generated image pairs $\{X^*, Y^*\}$ as the representative of old classes can be added to the dataset of the current session. We thus form a new training set on the current session by:

\begin{equation}
    \mathcal{D}^i_{train^*} = \mathcal{D}^i_{train} \cup \{X^*, Y^*\}
\end{equation}

Then, we can sample data pair $\{x, y\}$ from $\mathcal{D}^i_{train^*}$ to train current model $\mathcal{T}_i(\cdot;\theta_i)$ using cross-entropy $\mathcal{L}_{CE}(s,y)$, where s denotes the final probability vector given by cosine classifier \cite{ncm}. 
Note that re-labeling uncertain data with one-hot-like labels does not change the fact that the uncertain data is hard to be classified and important for transferring knowledge from an old model to a new one. It is shown in \cite{hard1,hard2} that using uncertain data (\textit{hard data}) can improve training efficiency.

It is also worth mentioning that most of the current methods \cite{zhu2021self,cheraghian2021synthesized} conduct incremental learning on a fixed backbone $\theta^b_{0}$ that is trained on the initial session. The trivial solution is sometimes beneficial because updating the backbone parameters in FSCIL can let the network easily over-fit on a few examples and forget the old mapping. However, keeping the backbone fixed can usually lead to defective model generalization on novel classes, as the backbone cannot provide discriminative features to novel classes. Thanks to our data replay, with the availability of both replayed data and novel data, we can fine-tune all the old parameters $\{\theta^b_{i-1}, \theta^{l}_{i-1} \}$ and update new task-specific parameters $\{\theta^{l^*}_i\}$ using:
\begin{equation}
    \begin{aligned}
    \{\theta^b_{i-1}, \theta^{l}_{i-1}\} &= \{\theta^b_{i-1}, \theta^{l}_{i-1}\} - \lambda_1  \frac{\partial \mathcal{L}_{CE}(s, y)}{\partial \{\theta^b_{i-1}, \theta^{l}_{i-1}\}} \\
    \theta^{l^*}_i &= \theta^{l^*}_i - \lambda_2  \frac{\partial \mathcal{L}_{CE}(s, y)}{\partial \theta^{l^*}_i}
    \end{aligned}
\end{equation}
where $\lambda_1$ and $\lambda_2$ denotes the learning rates for fine-tuning old parameters and updating new parameters.  After training, the new model of current session is parameterized by $\theta_i = \{\theta^b_i,\theta^l_i \}$, where $\theta^b_i = \theta^b_{i-1}$ and $\theta^{l}_i = \{\theta^{l}_{i-1}\, \theta^{l^*}_i\}$.
Due to the page limit, the overall algorithm is shown in the Appendix.

\section{Experiments}


In this section, we show the performance and several properties of our method through extensive experiments.

\subsection{Datasets}
We follow TOPIC \cite{topic} to conduct experiments on three widely used datasets including CIFAR-100 \cite{cifar100}, miniImageNet \cite{miniimagenet} and CUB-200 \cite{cub200}.

\noindent
\textbf{CIFAR-100} consists of 60,000 $32\times32$ color images in 100 classes, where the base training set contains 60 classes and incremental datasets are the collection of images from the remaining 40 classes. Each incremental dataset includes a total of 5 classes where each class has 5 training images.  

\noindent
\textbf{MiniImageNet} contains 100 classes that are sampled from the ILSVRC-12 dataset \cite{miniimagenet}. All images are $84 \times 84$ color images. The first 60 classes form the base dataset, and the remaining 40 classes are divided into 8 incremental sessions. Each incremental session is a 5-way 5-shot task.  

\noindent
\textbf{CUB-200} is a classification dataset containing  200 bird species. There are a total of 11,788 images for 200 classes, where the first 100 classes form our base dataset and the remaining 100 classes form 10 incremental sessions. In each incremental session, each class has 10 images with a resolution of $224\times224$.

\begin{table*}[!t]
\small
\centering
\caption{Comparison with the state-of-the-art methods on CIFAR-100, miniImageNet and CUB-200 datasets. Top rows are the CIL methods, and bottom rows are the FSCIL methods. * indicates our implementation of the methods under the FSCIL setting. + indicates the results reported in TOPIC \cite{topic}. Other results are copied
from the corresponding papers.}

\resizebox{11cm}{!} 
{ 
\begin{tabular}{c|lccccccccccc}

\Xhline{1pt}
\multirow{10}{*}{\rotatebox{90}{CIFAR-100}}&\multirow{2}{*}{Methods} &
\multicolumn{9}{c}{Sessions} & Average & Final \\
\cline{3-11} 
&& 0 & 1 & 2 & 3 & 4 & 5 & 6 & 7 & 8 & Acc & Impro. \\
\cline{2-13} 
&iCaRL+ \cite{icarl}  & 64.10& 53.28& 41.69& 34.13& 27.93& 25.06& 20.41& 15.48& 13.73& 32.87&+36.41\\
&iCaRL* \cite{icarl}  & 74.4& 67.67& 63.26& 59.68& 56.29& 53.48& 50.93& 48.76& 46.37& 57.87&+2.9\\
&NCM+ \cite{ncm} & 64.10& 53.05& 43.96& 36.97& 31.61& 26.73& 21.23& 16.78& 13.54& 34.22& +36.60\\

&NCM* \cite{ncm} & 74.4& 66.9& 64.13& 60.25& 58.37& 54.87& 51.74& 49.53& 47.21& 58.60& +2.17\\
&DeepInv* \cite{yin2020dreaming} & 74.4& 68.22& 64.37& 59.09& 56.87& 52.26& 50.77& 48.21& 47.08& 57.92& +2.85 \\
\cline{2-13}
&TOPIC+ \cite{topic} & 64.10& 55.88& 47.07& 45.16& 40.11& 36.38& 33.96& 31.55& 29.37& 42.62& +20.77\\
&Zhu et al.\cite{zhu2021self}  & 64.10& 65.86& 61.36& 57.34& 53.69& 50.75& 48.58& 45.66& 43.25& 54.51& +6.89\\
&Cheraghian et al.\cite{cheraghian2021synthesized}  & 62.00& 57.00& 56.7& 52.00& 50.60& 48.8& 45.00& 44.00& 41.64& 50.86 & +8.5\\
&CEC \cite{cec}  & 73.07& 68.88& 65.26& 61.19& 58.09& 55.57& 53.22& 51.34& 49.14& 59.53& +1.00\\

\cline{2-13}
&Ours & \textbf{74.4}& \textbf{70.2}& \textbf{66.54}& \textbf{62.51}& \textbf{59.71}& \textbf{56.58}& \textbf{54.52}& \textbf{52.39}& \textbf{50.14}& \textbf{60.77}&- \\
\Xhline{1pt}
\end{tabular}}

\vspace{1em}

\resizebox{11cm}{!} 
{ 
\begin{tabular}{c|lccccccccccc}
\Xhline{1pt}
\multirow{10}{*}{\rotatebox{90}{miniImageNet}}&\multirow{2}{*}{Methods} &
\multicolumn{9}{c}{Sessions} & Average & Final \\
\cline{3-11} 
&&0 & 1 & 2 & 3 & 4 & 5 & 6 & 7 & 8 & Acc & Impro. \\
\cline{2-13}
&iCaRL+ \cite{icarl}  & 61.31& 46.32& 42.49& 37.63& 30.49& 24.00& 20.89& 18.80& 17.21& 33.24&+31.0\\
&iCaRL* \cite{icarl}  & 71.84& 63.82& 59.43& 56.88& 53.14& 50.06& 48.37& 45.89& 44.13& 54.84&+3.18\\
&NCM+ \cite{ncm} & 61.31& 47.80& 39.31& 31.91& 25.68& 21.35& 18.67& 17.24& 14.17& 30.83& +34.04\\
&NCM* \cite{ncm} & 71.84& 66.52& 62.18& 57.93& 54.02& 50.89& 47.26& 45.83& 42.36& 55.43& +2.59\\
&DeepInv* \cite{yin2020dreaming} & 71.84&64.87& 61.43& 58.46& 56.62 &52.21& 49.42&47.26&45.06& 56.35& +1.67 \\
\cline{2-13}
&TOPIC+ \cite{topic} & 61.31 &50.09 &45.17 &41.16 &37.48 &35.52 &32.19 &29.46 &24.42 & 39.64& +23.79\\
&Zhu et al.\cite{zhu2021self}  & 61.45& 63.80& 59.53& 55.53& 52.50& 49.60& 46.69& 43.79& 41.92& 52.75& +6.29\\
&Cheraghian et al.\cite{cheraghian2021synthesized}  & 61.40& 59.80& 54.20& 51.69& 49.45& 48.00& 45.20& 43.80& 42.1& 50.63& +6.11\\
&CEC \cite{cec} & 72.00& 66.83& 62.97& 59.43& 56.70& 53.73& 51.19& 49.24& 47.63&  57.75& +0.58\\
\cline{2-13}
&Ours  & \textbf{71.84}& \textbf{67.12}& \textbf{63.21}& \textbf{59.77}& \textbf{57.01}& \textbf{53.95}& \textbf{51.55}& \textbf{49.52}& \textbf{48.21}& \textbf{58.02}&-  \\
\Xhline{1pt}
\end{tabular}}

\vspace{1em}

\resizebox{12cm}{!} 
{ 
\begin{tabular}{l|lccccccccccccccc}

\Xhline{1pt}
\multirow{10}{*}{\rotatebox{90}{CUB-200}}&\multirow{2}{*}{Methods} &
\multicolumn{11}{c}{Sessions} & Average & Final \\
\cline{3-13} 
&& 0 & 1 & 2 & 3 & 4 & 5 & 6 & 7 & 8 & 9 & 10 & Acc & Impro. \\
\cline{2-15}
&iCaRL+ \cite{icarl} & 68.68& 52.65& 48.61& 44.16& 36.62& 29.52& 27.83& 26.26& 24.01& 23.89& 21.16&36.67 &+31.23\\
&iCaRL* \cite{icarl} & 75.9& 63.32& 60.08 & 57.89& 53.89& 51.76& 48.88& 47.76& 44.92& 43.18& 41.37&53.54 &+7.98\\
&NCM+ \cite{ncm} & 68.68& 57.12& 44.21& 28.78& 26.71& 25.66& 24.62& 21.52& 20.12& 20.06& 19.87& 32.49&+32.52\\
&NCM* \cite{ncm} & 75.9& 65.89& 57.73& 52.08& 48.36& 43.38& 39.59& 36.02& 33.68& 32.01& 30.87& 46.86&+14.66\\
&DeepInv* \cite{yin2020dreaming}  & 75.90& 70.21&65.36&60.14&58.79&55.88& 53.21&51.27& 49.38&47.11&45.67& 57.54&+3.98\\
\cline{2-15}
&TOPIC+ \cite{topic} & 68.68& 62.49& 54.81& 49.99& 45.25& 41.40& 38.35& 35.36& 32.22& 28.31& 26.28& 43.92&+17.58\\
&Zhu et al.\cite{zhu2021self}  & 68.68& 61.85& 57.43& 52.68& 50.19& 46.88& 44.65& 43.07& 40.17& 39.63& 37.33& 49.32&+15.06\\
&Cheraghian et al.\cite{cheraghian2021synthesized}  & 68.78& 59.37& 59.32& 54.96& 52.58& 49.81& 48.09& 46.32& 44.33& 43.43& 43.23& 51.84& +9.16\\
&CEC \cite{cec} & 75.85& 71.94& 68.50& 63.50& 62.43& 58.27& 57.73& 55.81& 54.83& 53.52& 52.28& 61.33&+0.11\\
\cline{2-15}
&Ours  & \textbf{75.90}& \textbf{72.14}& \textbf{68.64}& \textbf{63.76}& \textbf{62.58}& \textbf{59.11}& \textbf{57.82}& \textbf{55.89}& \textbf{54.92}& \textbf{53.58}& \textbf{52.39}& \textbf{61.52} &-\\
\Xhline{1pt}
\end{tabular}}
\label{tab:compare_full_sota}
\end{table*}

\subsection{Implementation Details}

\subsubsection{Backbone network. } We implement our classification network by selecting from the off-the-shelf structures following TOPIC \cite{topic}. Specifically, we adopt ResNet20 \cite{he2016deep} as the backbone for the experiments on CIFAR-100 and use ResNet18 \cite{he2016deep} for the experiments on miniImageNet and CUB-200. We use the same network structure for the main network $\mathcal{T}$ and the auxiliary network $\mathcal{A}$. The generator of DCGAN \cite{dcgan} is adopted for replaying observed data in all the experiments.

\subsubsection{Training details. }
For all the experiments, we follow \cite{topic} to conduct training on base classes for 100 epochs using SGD with momentum and a batch size of 128. The learning rate is initialized to 0.1 with decay by a factor of 0.1 at epoch 60 and 70. We train the generator for 300 epochs at the beginning of each incremental session using Adam optimizer. The initial learning rate for updating the auxiliary network and the generator is set to 0.001 and 0.1, respectively. Learning rate decay is used on epochs 100, 150, and 200 by a factor of 0.1 for the auxiliary network and the generator. We then conduct incremental learning on a new model initialized by the old one using both the generated data and novel data. The quantity of generated data is the same as the number of training samples for all experiments. The learning rate $\lambda_1$ of backbone and old classifier is set to 0.0001, $\lambda_2 = 0.1$ for new nodes. Each incremental learning stage lasts for 40 epochs, with the learning rate decayed at $10$-th and $30$-th epochs. Data augmentations, such as random crop,  random scale, and random flip, are adopted at training time.

\subsection{Re-implementation of Replay-based Methods.}
To confirm our intuition that data replay can be used in FSCIL, we re-implement two replay-based approaches (i.e., iCaRL\cite{icarl} and NCM \cite{ncm}) under our setting. Table \ref{tab:compare_full_sota} presents the comparison between our obtained results (marked with *) and that reported in TOPIC \cite{topic} (marked with +).   The comparison shows that the performance of replay-based CIL methods is actually competitive under the FSCIL setting. The two methods outperform several state-of-the-arts by a significant margin. For example, the NCM \cite{ncm} outperform TOPIC \cite{topic}, zhu et al. \cite{zhu2021self}, and Cheraghian et al. \cite{cheraghian2021synthesized} by 17.71\%, 3.83\% and 5.44\% on CIFAR-100.  This fact indicates that data replay is indeed preferred and can potentially be adopted to address the FSCIL problem.

\subsection{Main Results and Comparison}
We conduct comparison with several methods, including three replay-based methods (i.e., iCaRL\cite{icarl}, NCM \cite{ncm} and DeepInversion (DeepInv) \cite{yin2020dreaming}) and four FSCIL methods (i.e., TOPIC \cite{topic}, Zhu et.al\cite{zhu2021self}, Cheraghian et.al\cite{cheraghian2021synthesized} and CEC \cite{cec}) . 
Table \ref{tab:compare_full_sota} summarizes the top-1 accuracy and average accuracy on all three benchmarks. Our main observations are as follows.
\begin{itemize}
\item Our method outperforms all the state-of-the-art on all three benchmarks across all sessions. The comparisons with the most recent method  \cite{cheraghian2021synthesized} illustrate the superiority of our proposed one. Specifically, our method achieves a final accuracy improvement over Cheraghian et al. \cite{cheraghian2021synthesized} by 8.5\%, 6.11\% and 9.16\% on CIFAR-100, miniImageNet and CUB-200, respectively. 
We outperform the existing state-of-the-art method (CEC \cite{cec}) by 1.00\%, 0.58\%, and 0.11\%. The superior performance of our approach further proves the effectiveness of adopting data replay in handling FSCIL problem.
\item Despite replay-based methods performing well, we still outperform them by a large margin.  
Since all the replayed-based methods are proposed to handle class-incremental learning and not specifically designed to address FSCIL, the performance of them is somewhat limited. In contrast, our method can successfully incorporate data replay in FSCIL.
\end{itemize}

\begin{table*}[t!]
\footnotesize
\centering

\setlength{\tabcolsep}{2pt} 

\caption{Ablation studies on CIFAR-100. We research the effects of adopting \textit{entropy regularization} (ER), \textit{re-labeling} (RL) and \textit{backbone fine-tuning} (BF).}
\begin{tabular}{ccccccccccccc}
\Xhline{1pt}
\multirow{2}{*}{ER}&\multirow{2}{*}{RL}&\multirow{2}{*}{BF} &
\multicolumn{9}{c}{Sessions} & Average  \\
\cline{4-12} 
& & &0 & 1 & 2 & 3 & 4 & 5 & 6 & 7 & 8 & Acc  \\
\cline{1-13} 
& \checkmark & \checkmark  & 74.4& 70.02& 66.24& 62.27& 59.38& 56.01& 53.76& 51.78& 49.50& 60.39\\
\checkmark& & \checkmark  & 74.4& 69.32& 65.78& 61.65& 58.32& 55.27& 53.19& 50.88& 48.92& 59.74\\
\checkmark&\checkmark &  & 74.4& 69.45& 65.92& 61.59& 58.66& 55.53& 53.24& 51.14& 49.13& 59.90 \\
\checkmark& \checkmark&\checkmark  &\textbf{74.4}& \textbf{70.2}& \textbf{66.54}& \textbf{62.51}& \textbf{59.71}& \textbf{56.58}& \textbf{54.52}& \textbf{52.39}& \textbf{50.14}& \textbf{60.77}\\
\Xhline{1pt}
\end{tabular}

\label{tab:ablation}
\end{table*}

\subsection{Analysis}

\subsubsection{Importance of different components.}
We firstly conduct ablation studies to reveal the effectiveness of different components in our proposed method. To be specific, we analyze the effects of \textit{entropy regularization} (ER) in Section \ref{sec:entropy}, re-labeling (RL) in Section \ref{sec:incremental} and backbone fine-tuning (BF) in Section \ref{sec:incremental}. The results are illustrated in Table \ref{tab:ablation}. It can be observed that the method with all the proposed components outperforms the others. By further comparing the full solution to the methods that are removed by one component, we can have several observations. 1) By removing the entropy regularization (shown in the first row in Table \ref{tab:ablation}), we notice that the final accuracy decreases. This shows that it is desirable to generate uncertain data for replay. 2) By removing the re-labeling of generated data and alternately following the vanilla knowledge distillation using KL divergence as distillation loss, we can observe from the second row of Table \ref{tab:ablation} that the performance drops significantly. This indicates the original knowledge distillation method is not appropriate for solving the specific problem. 3) We can see the benefit from updating our backbone parameters. Usually, simply updating the parameters of the backbone network is harmful, as the backbone can quickly over-fit on the few samples during incremental learning. Our data-free replay naturally inherits the advantage of fine-tuning, which allows a model to learn new classes as well as avoid forgetting.
Thus, we can observe the improvements of adopting backbone fine-tuning by comparing the last two rows in Table \ref{tab:ablation}.

\begin{figure}[t]
    \centering
    \subfigure[]{%
        \includegraphics[width=0.4\linewidth]{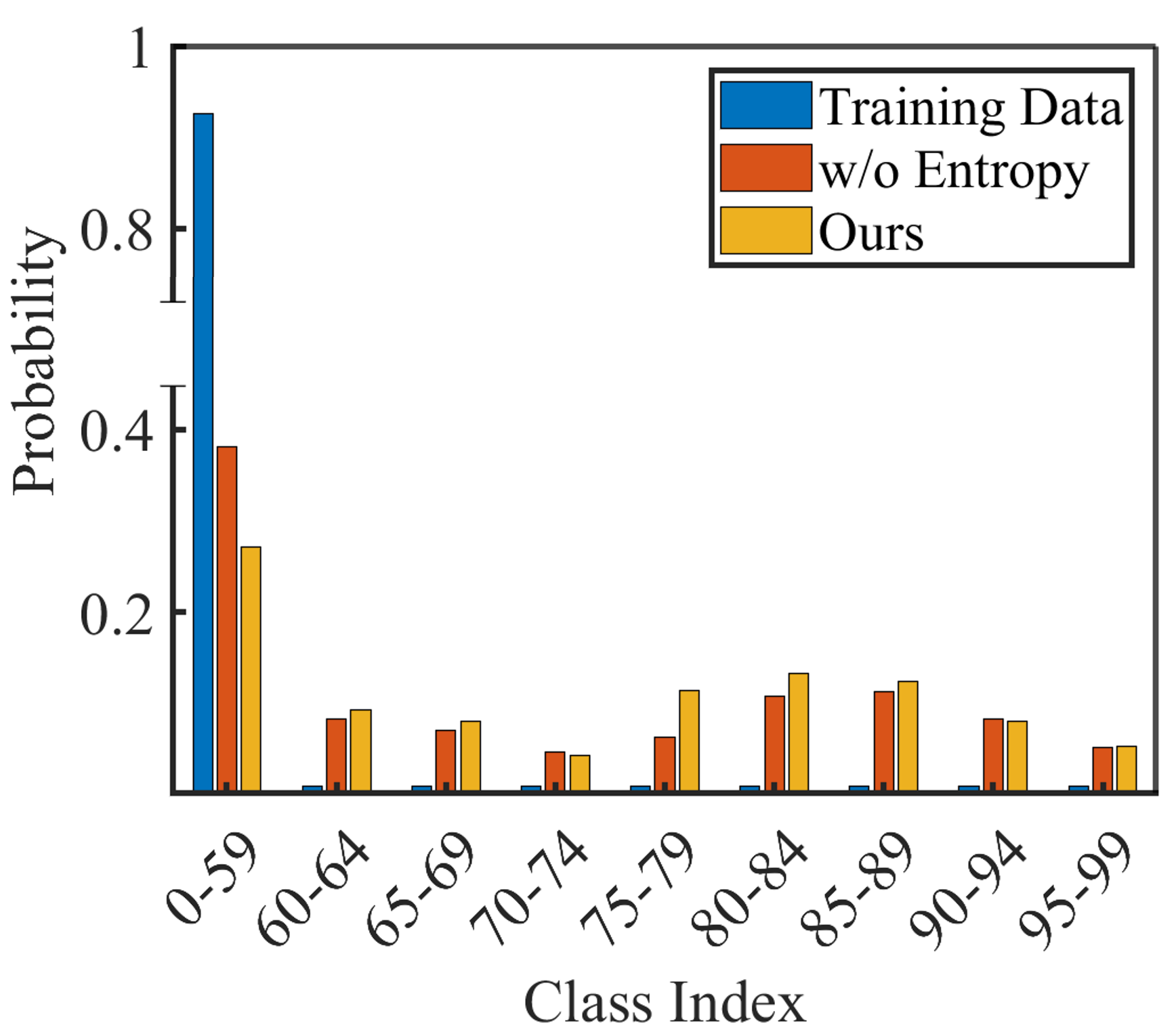}
        \label{fig:label_distribution}
    } 
    \subfigure[]{\raisebox{3.2mm}{
        \includegraphics[width=0.418\linewidth]{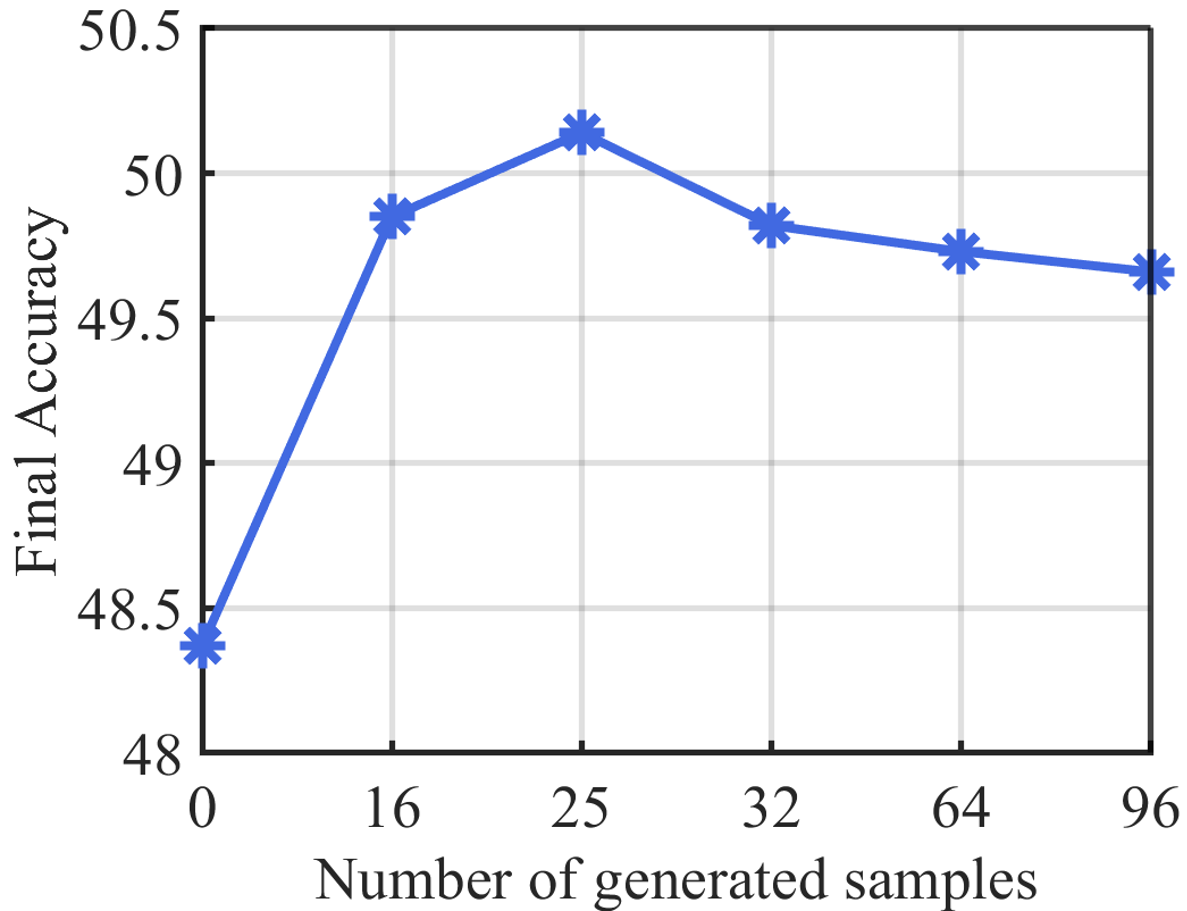}}
        \label{fig:num}
    } 
	\caption{ \subref{fig:toy_wo_entropy} shows the label distribution of training data and generated data. \subref{fig:num} shows the average accuracy of the proposed method given different number of replayed data per batch. }
	\label{fig:toy}
\end{figure}

\subsubsection{Label distribution of generated data.} As is denoted in Sec \ref{sec:entropy}, our proposed entropy regularizer can encourage the generator to synthesize much data of 
 few-shot learned classes. However, since the incremental classes are trained using very limited data samples, it is somewhat doubtful that the generator can synthesize sufficient data for the new classes. To address this concern and further disclose
the property of our approach, we show in Figure \ref{fig:label_distribution} to illustrate the label distribution of the real data and our generated data. Besides, we also include the results produced by the generator without using our proposed entropy regularizer. It can be observed that using entropy regularizer can encourage the generator to synthesize more samples belonging to the few-shot learned classes. In contrast,  the generator trained without entropy regularizer tends to synthesize more images belonging to base classes.

\subsubsection{Analysis of number of generated data.} 
Figure \ref{fig:num} shows the impact 
of 
 adopting different numbers of generated samples in our proposed method. 
It illustrates that the performance of our method is maximized when the quantity of replayed data is the same as the training data (in CIFAR-100, each incremental session   provides 25 training images.).
 Using smaller or larger quantities of the replayed data can jeopardize the overall performance mainly because of the adverse effects of the imbalance \cite{ncm}. 
 A small number of replayed data might not be  
sufficient against forgetting, while too much replayed old data can potentially hinder
the network from adapting to new classes.  
 Note that we do not fix the input noise vectors across batches. This allows the generator to synthesize more diverse data samples of the observed classes.

\subsubsection{Analysis of data-free replay. }
In addition to the adopted method \cite{belief}, DeepInversion \cite{yin2020dreaming} provides an alternative solution to data-free replay. Here, we study the effect of replacing our adopted replay method with DeepInversion.  
\begin{wraptable}{l}{6.9cm}
\centering
\resizebox{7cm}{!} 
{ 
\begin{tabular}{ccccc}
\Xhline{1pt}
Method  &Session 0 & Session 8 & Average Acc & Final Impro.  \\
\hline
DeepInv&74.4&47.08&57.92&+2.85\\
DeepInv+Ours& 74.4& 48.48& 59.11 & +1.66\\
Ours & 74.4 & 50.14 & 60.77 & -\\
\Xhline{1pt}
\end{tabular}}
\caption{The impact of adopting DeepInversion (DI) as data-free replay method.}
\label{tab:deepinversion}
\end{wraptable} 
Table \ref{tab:deepinversion} summarizes the results. First, by comparing the first row with the second row, we can observe that using DeepInversion with our re-labeling and backbone fine-tuning can significantly boost its performance. Second, the comparison of the last two rows illustrates that our data replay approach is more suitable for FSCIL. Specifically, 
our method outperforms the case using DeepInversion by 1.66\% in terms of final accuracy. The potential issue of DeepInversion is that it encourages the generated data to have confident predictions by the old model. Due to the fact that certain predictions are relatively rare under the few-shot setting, the generated data might be ineffective.
Another issue we have observed for DeepInversion is that the data generation process is time-consuming.This is also confirmed in \cite{yin2020dreaming}.
In contrast, our generator can replay data very efficiently once it is trained to be ready.

\section{Conclusion}
In this paper,  we first disclose that data replay can be adopted in addressing FSCIL problem. Then we propose a novel approach to denote the effectiveness of using data replay in FSCIL. To address the privacy concern of vanilla data replay, we introduce the data-free replay scheme for synthesizing old samples. By observing that the prediction of the classification model becomes uncertain under the few-shot incremental setting, we propose an entropy regularization on the training of the generator. We then design a new method to learn from the uncertain data via re-labeling against forgetting issues. Extensive comparison with the state-of-the-arts illustrates that our approach achieves the best performance for avoiding forgetting and quickly adapting to new classes. 

\noindent
\textbf{Limitation.} 
Difficulties might occur when training the generator on large-scaled datasets, potentially jeopardizing replayed data quality.

\clearpage
%
%
\bibliographystyle{splncs04}
\bibliography{egbib}

\begin{thebibliography}{10}
\providecommand{\url}[1]{\texttt{#1}}
\providecommand{\urlprefix}{URL }
\providecommand{\doi}[1]{https://doi.org/#1}

\bibitem{rep1}
Caron, M., Misra, I., Mairal, J., Goyal, P., Bojanowski, P., Joulin, A.:
  Unsupervised learning of visual features by contrasting cluster assignments.
  Advances in Neural Information Processing Systems  \textbf{33},  9912--9924
  (2020)

\bibitem{rep2}
Caron, M., Touvron, H., Misra, I., J{\'e}gou, H., Mairal, J., Bojanowski, P.,
  Joulin, A.: Emerging properties in self-supervised vision transformers. In:
  Proceedings of the IEEE/CVF International Conference on Computer Vision. pp.
  9650--9660 (2021)

\bibitem{eeil}
Castro, F.M., Mar{\'\i}n-Jim{\'e}nez, M.J., Guil, N., Schmid, C., Alahari, K.:
  End-to-end incremental learning. In: European Confererence on Computer Vison
  (2018)

\bibitem{Data-free-student}
Chen, H., Wang, Y., Xu, C., Yang, Z., Liu, C., Shi, B., Xu, C., Xu, C., Tian,
  Q.: Data-free learning of student networks. In: IEEE International Conference
  on Computer Vision (2019)

\bibitem{hard1}
Chen, K., Chen, Y., Han, C., Sang, N., Gao, C.: Hard sample mining makes person
  re-identification more efficient and accurate. Neurocomputing  \textbf{382},
  259--267 (2020)

\bibitem{rep3}
Chen, T., Kornblith, S., Norouzi, M., Hinton, G.: A simple framework for
  contrastive learning of visual representations. In: International conference
  on machine learning. pp. 1597--1607. PMLR (2020)

\bibitem{rep4}
Chen, X., Fan, H., Girshick, R., He, K.: Improved baselines with momentum
  contrastive learning. arXiv preprint arXiv:2003.04297  (2020)

\bibitem{cheraghian2021synthesized}
Cheraghian, A., Rahman, S., Ramasinghe, S., Fang, P., Simon, C., Petersson, L.,
  Harandi, M.: Synthesized feature based few-shot class-incremental learning on
  a mixture of subspaces. In: IEEE International Conference on Computer Vision
  (2021)

\bibitem{metafscil}
Chi, Z., Gu, L., Liu, H., Wang, Y., Yu, Y., Tang, J.: Metafscil: A
  meta-learning approach for few-shot class incremental learning. In: IEEE/CVF
  Conference on Computer Vision and Pattern Recognition. pp. 14166--14175
  (2022)

\bibitem{gan_memory}
Cong, Y., Zhao, M., Li, J., Wang, S., Carin, L.: Gan memory with no forgetting.
  Advances in Neural Information Processing Systems  (2020)

\bibitem{hard2}
Felzenszwalb, P.F., Girshick, R.B., McAllester, D., Ramanan, D.: Object
  detection with discriminatively trained part-based models. IEEE Transactions
  on Pattern Analysis and Machine Intelligence  \textbf{32}(9),  1627--1645
  (2009)

\bibitem{fini2022self}
Fini, E., da~Costa, V.G.T., Alameda-Pineda, X., Ricci, E., Alahari, K., Mairal,
  J.: Self-supervised models are continual learners. In: Proceedings of the
  IEEE/CVF Conference on Computer Vision and Pattern Recognition. pp.
  9621--9630 (2022)

\bibitem{french1999catastrophic}
French, R.M.: Catastrophic forgetting in connectionist networks. Trends in
  Cognitive Sciences  \textbf{3}(4),  128--135 (1999)

\bibitem{he2016deep}
He, K., Zhang, X., Ren, S., Sun, J.: Deep residual learning for image
  recognition. In: IEEE/CVF Conference on Computer Vision and Pattern
  Recognition (2016)

\bibitem{ncm}
Hou, S., Pan, X., Loy, C.C., Wang, Z., Lin, D.: Learning a unified classifier
  incrementally via rebalancing. In: IEEE/CVF Conference on Computer Vision and
  Pattern Recognition (2019)

\bibitem{kl_mse}
Kim, T., Oh, J., Kim, N.Y., Cho, S., Yun, S.Y.: Comparing kullback-leibler
  divergence and mean squared error loss in knowledge distillation. In:
  International Joint Conference on Artificial Intelligence (2021)

\bibitem{cifar100}
Krizhevsky, A., Hinton, G., et~al.: Learning multiple layers of features from
  tiny images  (2009)

\bibitem{lecun2015deep}
LeCun, Y., Bengio, Y., Hinton, G.: Deep learning. Nature  \textbf{521}(7553),
  436--444 (2015)

\bibitem{liang2022self}
Liang, H., Quader, N., Chi, Z., Chen, L., Dai, P., Lu, J., Wang, Y.:
  Self-supervised spatiotemporal representation learning by exploiting video
  continuity. In: Proceedings of the AAAI Conference on Artificial
  Intelligence. vol.~36, pp. 1564--1573 (2022)

\bibitem{belief}
Micaelli, P., Storkey, A.J.: Zero-shot knowledge transfer via adversarial
  belief matching. In: Advances in Neural Information Processing Systems (2019)

\bibitem{dcgan}
Radford, A., Metz, L., Chintala, S.: Unsupervised representation learning with
  deep convolutional generative adversarial networks. International Conference
  on Learning Representations  (2016)

\bibitem{icarl}
Rebuffi, S.A., Kolesnikov, A., Sperl, G., Lampert, C.H.: icarl: Incremental
  classifier and representation learning. In: IEEE/CVF Conference on Computer
  Vision and Pattern Recognition (2017)

\bibitem{miniimagenet}
Russakovsky, O., Deng, J., Su, H., Krause, J., Satheesh, S., Ma, S., Huang, Z.,
  Karpathy, A., Khosla, A., Bernstein, M., et~al.: Imagenet large scale visual
  recognition challenge. International Journal of Computer Vision
  \textbf{115}(3),  211--252 (2015)

\bibitem{shankarampeta2021few}
Shankarampeta, A.R., Yamauchi, K.: Few-shot class incremental learning with
  generative feature replay. In: International Conference on Pattern
  Recognition Applications and Methods. pp. 259--267 (2021)

\bibitem{cl_gr}
Shin, H., Lee, J.K., Kim, J., Kim, J.: Continual learning with deep generative
  replay. In: Advances in Neural Information Processing Systems (2017)

\bibitem{smith2021always}
Smith, J., Hsu, Y.C., Balloch, J., Shen, Y., Jin, H., Kira, Z.: Always be
  dreaming: A new approach for data-free class-incremental learning. In: IEEE
  International Conference on Computer Vision (2021)

\bibitem{topic}
Tao, X., Hong, X., Chang, X., Dong, S., Wei, X., Gong, Y.: Few-shot
  class-incremental learning. In: IEEE/CVF Conference on Computer Vision and
  Pattern Recognition (2020)

\bibitem{cub200}
Wah, C., Branson, S., Welinder, P., Perona, P., Belongie, S.: The caltech-ucsd
  birds-200-2011 dataset  (2011)

\bibitem{xin2021memory}
Xin, X., Zhong, Y., Hou, Y., Wang, J., Zheng, L.: Memory-free generative replay
  for class-incremental learning. arXiv preprint arXiv:2109.00328  (2021)

\bibitem{yin2020dreaming}
Yin, H., Molchanov, P., Alvarez, J.M., Li, Z., Mallya, A., Hoiem, D., Jha,
  N.K., Kautz, J.: Dreaming to distill: Data-free knowledge transfer via
  deepinversion. In: IEEE/CVF Conference on Computer Vision and Pattern
  Recognition (2020)

\bibitem{cec}
Zhang, C., Song, N., Lin, G., Zheng, Y., Pan, P., Xu, Y.: Few-shot incremental
  learning with continually evolved classifiers. In: IEEE/CVF Conference on
  Computer Vision and Pattern Recognition (2021)

\bibitem{zhu2021self}
Zhu, K., Cao, Y., Zhai, W., Cheng, J., Zha, Z.J.: Self-promoted prototype
  refinement for few-shot class-incremental learning. In: IEEE/CVF Conference
  on Computer Vision and Pattern Recognition (2021)

\end{thebibliography}

\end{document}